\documentclass[preprint,12pt]{elsarticle}

\usepackage{amssymb}
\usepackage{amsmath}
\usepackage{graphicx}
\usepackage{booktabs}
\usepackage{multirow}
\usepackage{siunitx}
\usepackage[table]{xcolor}
\usepackage{xcolor}

\usepackage[normalem]{ulem}

\journal{Image and Vision Computing}

\begin{document}

\begin{frontmatter}



\title{DepthTCM: High Efficient Depth Compression via Physics-aware Transformer-CNN Mixed Architecture}


% \author{} %% Author name
\author[aff1]{Young-Seo Chang}
\ead{pdhnet2@yonsei.ac.kr}
\author[aff2]{Yatong An\corref{cor1}}
\ead{yatong@meta.com}
\author[aff1]{Jae-Sang Hyun\corref{cor1}}
\ead{hyun.jaesang@yonsei.ac.kr}
% Author affiliation

\affiliation[aff1]{
  organization={Department of Mechanical Engineering, Yonsei University},
  city={Seoul},
  postcode={03722},
  country={South Korea}
}

\affiliation[aff2]{
  organization={Meta Reality Labs},
  city={Redmond},
  state={WA},
  postcode={98052},
  country={USA}
}

\cortext[cor1]{Corresponding authors.}
%% Abstract
\begin{abstract}
We propose DepthTCM, a physics-aware end-to-end framework for depth map compression. In our framework of DepthTCM, the high-bit depth map is first converted to a conventional 3-channel image representation losslessly using a method inspired by a physical sinusoidal fringe pattern based profiliometry system, then the 3-channel color image is encoded and decoded by a recently developed Transformer-CNN mixed neural network architecture. Specifically, DepthTCM maps depth to a smooth 3-channel using multiwavelength depth (MWD) encoding, then globally quantized the MWD encoded representation to 4 bits per channel to reduce entropy, and finally is compressed using a learned codec that combines convolutional and Transformer layers. Experiment results demonstrate the advantage of our proposed method. On Middlebury 2014, DepthTCM reaches 0.307 bpp while preserving 99.38\% accuracy, a level of fidelity commensurate with lossless PNG. We additionally demonstrate practical efficiency and scalability, reporting average end-to-end inference times of 41.48 ms (encoder) and 47.45 ms (decoder) on the ScanNet++ iPhone RGB-D subset. Ablations validate our design choices: relative to 8-bit quantization, 4-bit quantization reduces bitrate by 66\% while maintaining comparable reconstruction quality, with only a marginal 0.68 dB PSNR change and a 0.04\% accuracy difference. In addition, Transformer--CNN blocks further improve PSNR by up to 0.75 dB over CNN-only architectures.
\end{abstract}

\begin{graphicalabstract}
\centering
  \includegraphics[width=\linewidth]{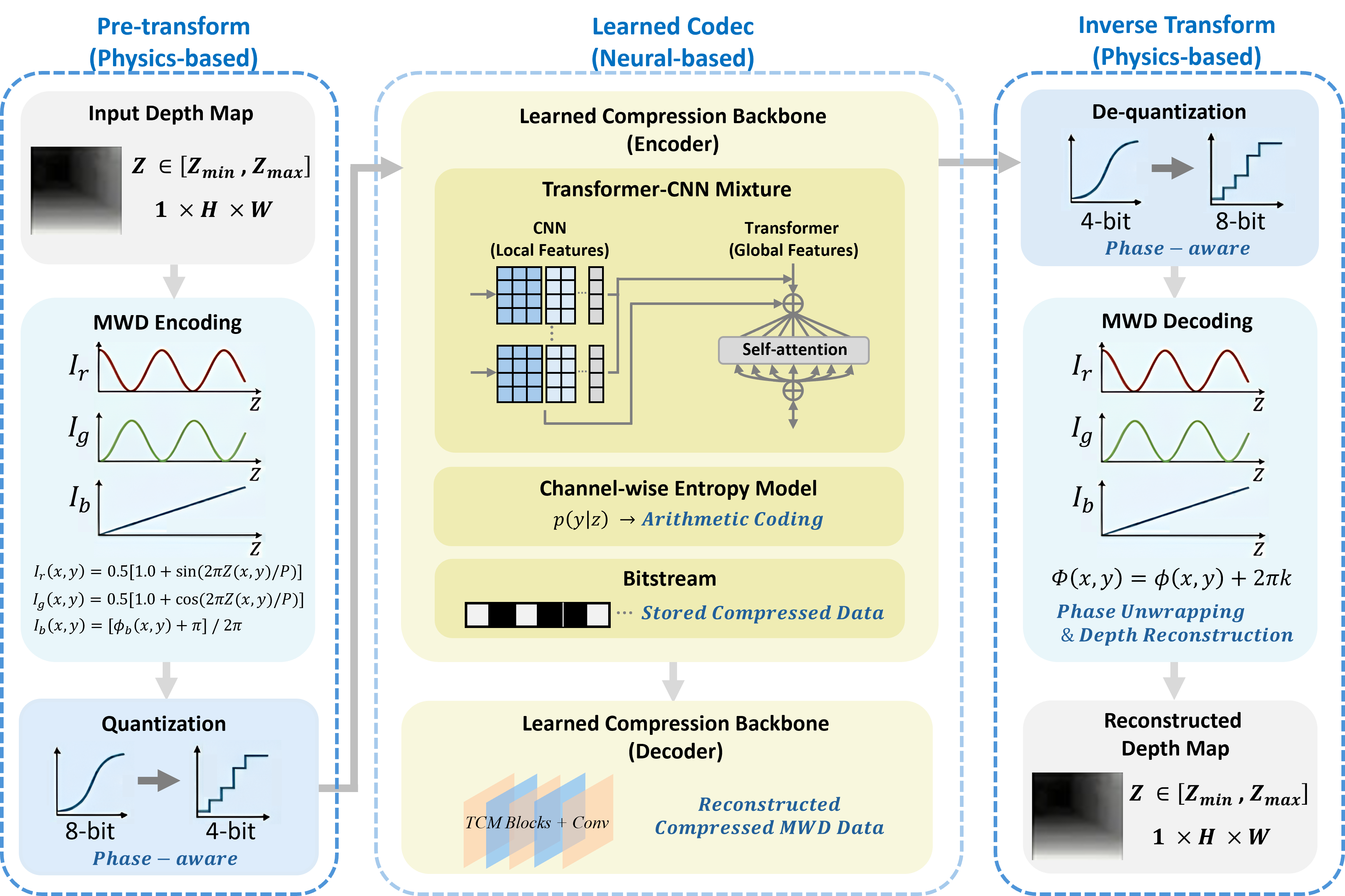}
\end{graphicalabstract}

\begin{highlights}
\item Modern learned image compression is effective for multiwavelength depth signals.
\item Sinusoidal MWD representations are quantized using a global 4-bit scheme.
\item Low-bit MWD quantization preserves geometric fidelity and improves entropy coding.
\item The end-to-end pipeline does not rely on auxiliary RGB information.
\item Consistent rate--distortion gains are observed across standard depth benchmarks.
\end{highlights}

%% Keywords

\end{frontmatter}



\section{Introduction}
\label{sec1}
%% Labels are used to cross-reference an item using \ref command.
\begin{figure}[t]
  \centering
  \includegraphics[width=1.0\linewidth]{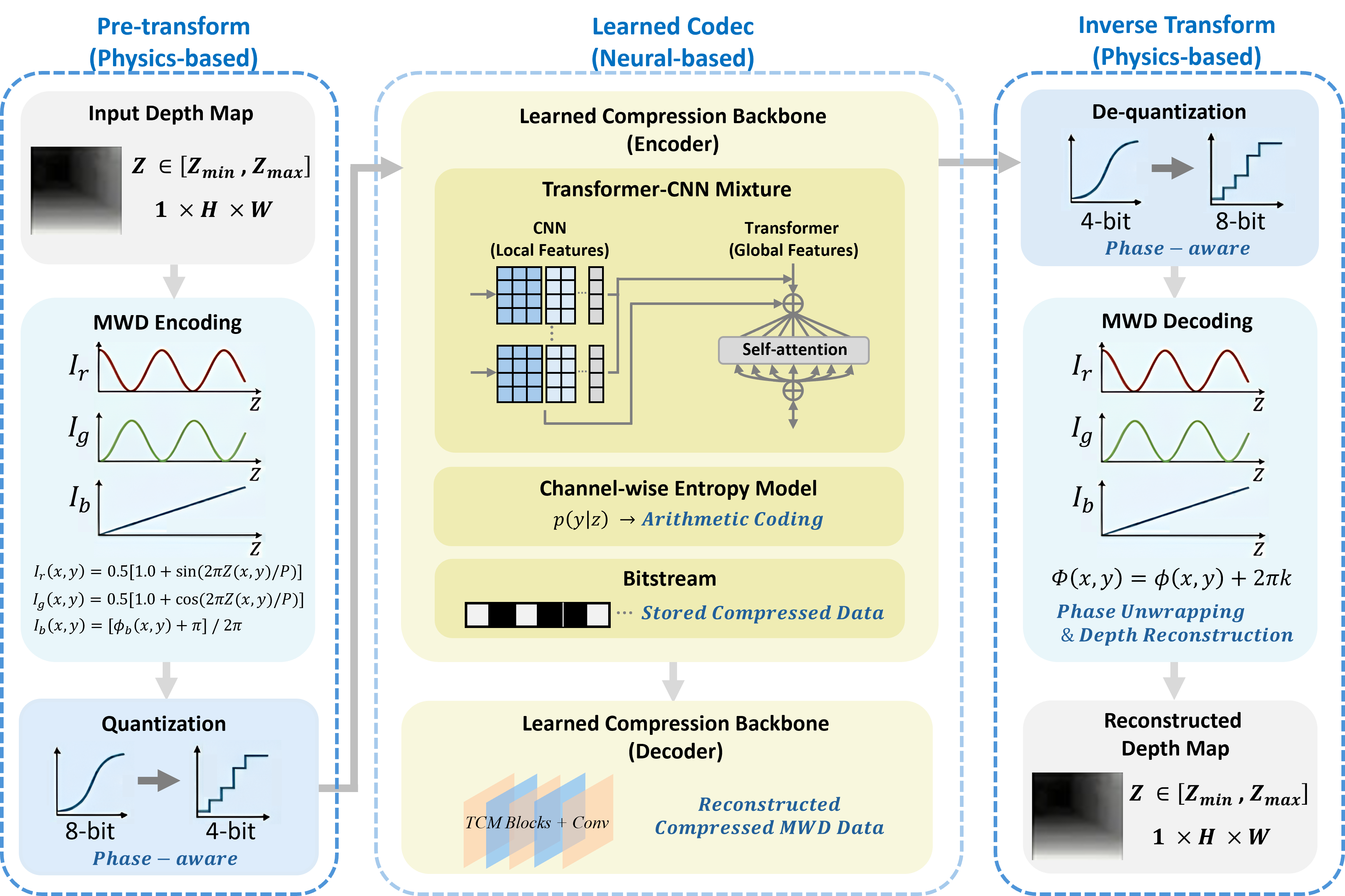}
    \caption{DepthTCM framework architecture. The framework integrates multiwavelength depth (MWD) encoding, 4-bit quantization, and a hybrid Transformer--CNN compression backbone into a fully differentiable end-to-end system for efficient depth map compression and reconstruction.}
   \label{fig:pipeline}
\end{figure}
Over the past several decades, the field of depth sensing and 3D reconstruction has witnessed substantial progress in both hardware and algorithmic developments, including innovations such as high-speed structured light systems \cite{zhang2018high}, LIDAR \cite{doi:10.1177/0278364913491297}, and novel-view rendering algorithms \cite{mildenhall2020nerf}. These advances have greatly contributed to the widespread availability of high-resolution and high-accurate 3D depth maps. However, high-quality depth maps offer precise geometric details, but this richness of information usually comes at the cost of large data volumes, necessitating efficient compression and real-time transmission strategies for storage and streaming in view-based rendering and telepresence applications. Image compression techniques can serve as valuable references for depth compression and have undergone extensive development over the last several decades. Numerous methods, including classical codecs such as JPEG\cite{125072} and PNG\cite{iso15948}, as well as recent machine learning based methods \cite{balle2018variational} have been widely adopted in practical applications or well established and thoroughly studied for natural images. However, those methods typically cannot be directly applied to depth data due to the distinct characteristics of depth maps, which consist of piecewise smooth surfaces bounded by sharp discontinuities \cite{6115989}. Consequently, conventional image codecs often fail to efficiently represent these high-frequency edges, leading to bloated bitrates or loss of critical structural details \cite{5702565}.

To tackle the above challenge, progresses have been made in various types of depth compression methods, including conventional ones without learning, learning based ones and hybrid ones. Conventional non-learning methods typically adapt classical image compression frameworks to accommodate depth map characteristics. Prominent examples include the depth modeling modes in 3D-HEVC \cite{7258339}, graph-based transform methods \cite{5702565}, and encoding schemes like MWD \cite{bell2015multiwavelength}. This type of methods are well explainable and easily adoptable for practical applications. For learning based methods, they typically apply recent deep learning developments in the 2D image processing field to the depth data. End-to-end learned codecs for depth images \cite{wu2022end} have shown significant advances by directly optimizing rate-distortion objectives on depth data. Recent end-to-end learned approaches provide variable-rate control via a single parameter $\lambda$ but commonly evaluate with synthesized-view RD (original texture combined with compressed depth), which complicates cross-dataset reproducibility and may not reflect depth-domain distortion directly \cite{sebai2024end}. Building on the ideas of non-learning and learning methods, hybrid methods try to combine merits of both types of methods. The idea of hybrid methods is usually change part of a conventional method to be learning based to optimize its performance. A notable example of hybrid methods is N-DEPTH \cite{siemonsma2024neural}, which uses neural networks to map depth to an RGB image optimized for standard codecs, and outperform MWD in compression fidelity.

Though great progresses have been made in hybrid methods, this type of methods have not unleashed its full potential yet. For instance, N-DEPTH \cite{siemonsma2024neural} still depend on traditional codecs (JPEG or H.264) for final bitstream encoding, that is not specifically optimized for depth modality. Recent advances in learned image compression (LIC) propose end-to-end neural codecs with sophisticated entropy models and Transformer architectures, offering substantial improvements in rate-distortion performance, making it is possible to learn a dedicated codec for MWD encoded depth images. However, directly applying modern learned image compression architectures to raw depth maps often yields suboptimal performance. Depth maps contain large homogeneous regions separated by sharp discontinuities at object boundaries, resulting in boundary-dominated signal statistics that differ significantly from natural image textures. This observation motivates the use of a physics-inspired pre-transform that reshapes depth statistics into a more compressible representation. Based on the above two developments and to unleash the potential of hybrid methods, in this paper, we propose DepthTCM, an end-to-end depth compression framework that integrates multiwavelength encoding with a state-of-the-art learned image codec.

The framework of DepthTCM is shown in Fig.\ref{fig:pipeline}. DepthTCM integrates a physics-based multiwavelength pre-transform with a differentiable learned codec and a phase-aware global 4-bit quantization, enabling depth-only, variable-rate compression that is tailored to the statistics of geometric signals and validated on standard depth benchmarks. It firstly leveraged the physics inspired MWD encoding method\cite{bell2015multiwavelength}, which encodes depth into three color channels using sinusoidal patterns. Then instead of using the classical JPEG to do compression, we trained end-to-end with a rate-distortion objective, DepthTCM reduces bitrate by 60.3\% compared to previous multiwavelength-based methods, while maintaining lossless-grade accuracy (over 99\%) at matched fidelity on both DIODE and Middlebury benchmarks. These results highlight the effectiveness of combining model-based depth encoding with learned compression for efficient and accurate geometric data representation.

DepthTCM demonstrates that a physics-inspired phase representation fundamentally reshapes the entropy characteristics of depth signals, enabling general-purpose learned image codecs to operate efficiently on geometric data. Our main contributions are summarized as follows:
\begin{itemize}
    \item A novel physics-inspired depth compression framework: We formulate a fully differentiable pipeline that enables joint optimization of physics-inspired MWD encoding together with a high-capacity learned compression codec.
    \item Phase-aware low-bit quantization while preserving geometric fidelity: We propose global 4-bit quantization to the MWD-encoded representation prior to compression, exploiting phase structure to reduce entropy while preserving geometric fidelity.
    \item High compression ratio while maintaining great accuracy: Extensive evaluation on standard benchmarks was conducted, from which we demonstrate substantial bitrate reduction while maintaining lossless-grade reconstruction accuracy (\textgreater 99\%) on DIODE and Middlebury benchmarks.
\end{itemize}

\section{Related work}
Depth image compression has seen rapid progress in recent years, spanning end-to-end learned approaches, hybrid neural-plus-codec techniques, and analytic methods tailored to depth data characteristics. Unlike natural RGB images, depth maps exhibit piecewise smooth surfaces with sharp discontinuities at object boundaries, so specialized compression strategies have emerged to exploit these properties.

\subsection{Analytical methods for depth compression}
Analytical methods usually tailor conventional image codec methods to be suitable for depth. There are two categories of methods: one directly perform compression on the depth data, and the other one will convert depth map to a different image based representation and do compression on the new representation. Example of the first category methods include tree-based methods \cite{sarkis2009fast,ogniewski2017best}, handcrafted geometric feature based methods \cite{gautier2012efficient}, which usually relies on geometric primitives to sparsify the data. The second category usually convert depth into another representation inspired by the depth generation principles. For instance, Fu et al. \cite{fu2013kinect} exploited the inherent depth–disparity relation in Kinect sensors to guide the compression design. Bell. etc encoded depth maps into pseudo-RGB images suitable for standard compression pipelines. Bell and Zhang \cite{bell2015multiwavelength} presented Multiwavelength Depth (MWD) encoding that represents depth as phase patterns across RGB channels, and Schwartz and Bell \cite{schwartz2022downsampled} further improve this with downsampling and upsampling strategies. 

\subsection{Learned Depth Compression}
End-to-end learned codecs for depth images have shown significant advances by directly optimizing rate-distortion objectives on depth data. Sebai et al. \cite{sebai2024end} propose a quality-scalable autoencoder that incorporates wedgelet edge filters and a depth-versus-texture classifier, enabling better preservation of depth discontinuities and outperforming 3D-HEVC. Wu and Gao \cite{wu2022end} achieve efficient lossless compression by cascading a learned lossy codec with a lossless coder, introducing pseudo-residuals to improve entropy modeling at low complexity. Implicit neural representation approaches, such as Ladune et al.'s COOL-CHIC \cite{ladune2023cool} and Kuwabara et al. \cite{kuwabara2025range}, fit compact MLPs per depth map to reach VVC-level rates and excel in 3D reconstruction at low bitrates. 

Recent learned image compression models for natural RGB images increasingly fuse convolutional networks with Transformer modules to capture both fine details and long-range dependencies. Qian et al.'s Entroformer \cite{qian2022entroformer} employs top-k attention for entropy modeling, while Contextformer and its efficient variant \cite{koyuncu2022contextformer, koyuncu2024efficient} replace masked convolutions with spatio-channel self-attention, achieving notable bitrate reductions and faster decoding. Hybrid autoencoders that interleave CNNs with window-based or parallel Transformer blocks \cite{liu2023learned, zou2022devil} further improve global structure preservation. DepthTCM builds on these advances by embedding Transformer-CNN hybrid blocks within its codec, enabling accurate edge reconstruction and robust geometric fidelity in depth maps.

\subsection{ Hybrid Approaches}

Although learning based methods leverage data-driven features for superior rate-distortion performance, they often neglect depth maps' trigonometric structure, undermining geometric fidelity under aggressive quantization. Therefore, another promising direction combines neural networks with conventional codecs by encoding depth maps into pseudo-RGB images suitable for standard compression pipelines. Bell and Zhang \cite{bell2015multiwavelength} presented Multiwavelength Depth (MWD) encoding that represents depth as phase patterns across RGB channels, and Schwartz and Bell \cite{schwartz2022downsampled} further improve this with downsampling and upsampling strategies. Siemonsma and Bell's N-DEPTH \cite{siemonsma2024neural} extends this by introducing a differentiable JPEG module within an end-to-end learned mapping, substantially lowering RMSE and file size compared to handcrafted MWD, especially under JPEG and H.264 compression. Siekkinen and Kämäräinen's network \cite{siekkinen2023neural} predicts optimal bit precision per frame to balance compression ratio and accuracy. These hybrid approaches leverage codec robustness and learning-based adaptation for depth-specific challenges, although they remain partly constrained by codec assumptions.

\section{Method}
\subsection{Overview}
Fig.~\ref{fig:pipeline} illustrates the overall DepthTCM framework architecture, consisting of both compression and decompression stages. Given a single-channel depth map, DepthTCM first transforms it into a smooth three-channel representation through multiwavelength depth (MWD) encoding and quantization, which is then processed by a Transformer-CNN compression backbone.

\textbf{Compression Pipeline:} (1) MWD encoding: Convert depth map $Z$ to an 8-bit RGB image via multiwavelength transform, where $P$ denotes the fringe period in depth units. The red channel is $\sin(2\pi Z/P)$, green is $\cos(2\pi Z/P)$, and blue is normalized depth value. (2) Quantization: Reduce the bit depth of the MWD image to 4-bit per channel. (3) Analysis transform: Process the quantized image through convolutional and Transformer-CNN mixture (TCM) blocks, generating latent codes $\mathbf{y}$ and hyperprior side latent information $\mathbf{z}$. (4) Entropy coding: Arithmetic-code $\mathbf{y}$ and $\mathbf{z}$ into the output bitstream. The bitstream includes compressed codes and side information (e.g., shape, transform parameters).

\textbf{Decompression Pipeline:} (1) Entropy decoding: Recover $\mathbf{y}$ and $\mathbf{z}$ from the bitstream using the learned entropy model. (2) Synthesis transform: Decode $\mathbf{y}$ through the TCM-based decoder to reconstruct the quantized MWD image. (3) De-quantization: Convert the 4-bit image to restore 8-bit range values. (4) MWD decoding: Apply the inverse multiwavelength decoding to reconstruct depth the final depth map $\hat{Z}$.

The entire framework is end-to-end differentiable, which allows for joint optimization of all components for the rate-distortion objective. Depth values are encoded using sinusoidal MWD patterns, which preserve smooth geometric variation and leverage cross-channel trigonometric constraints to mitigate quantization errors. This allows the network to accurately recover original depth maps even in the presence of channel perturbations.

\begin{figure}[t]
\centering
\includegraphics[width=1.0\textwidth]{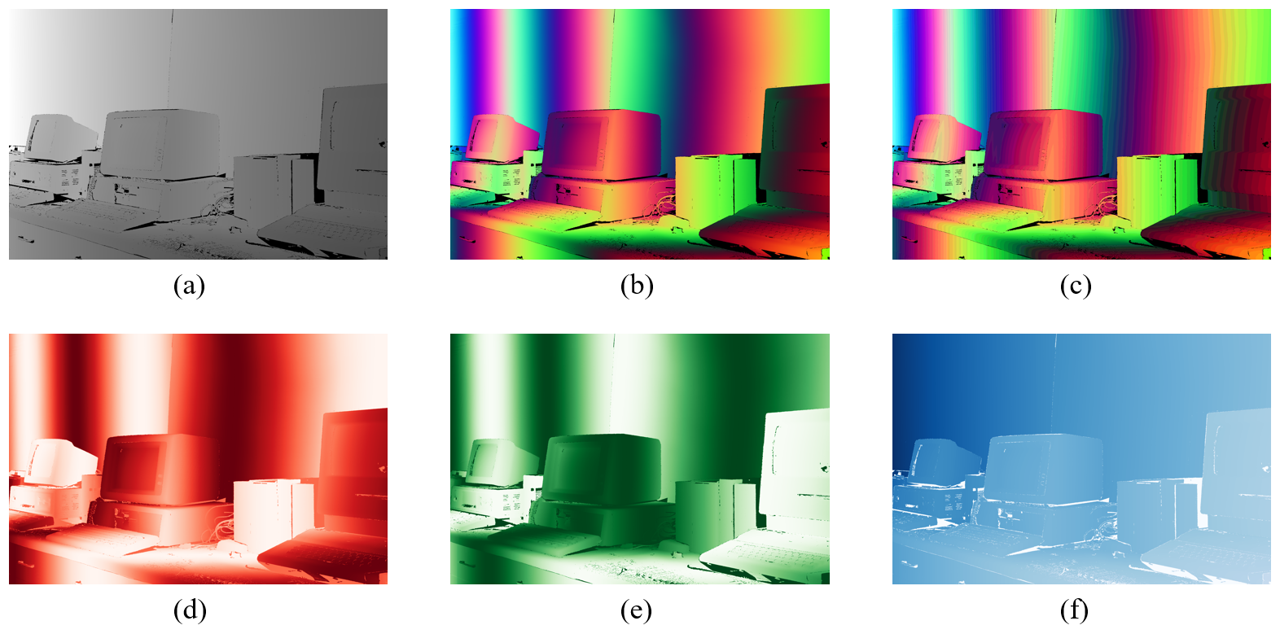}
\caption{Visualization of Multiwavelength Depth Encoding. (a) Input 32-bit depth map, (b) 8-bit Quantized MWD Encoded Map, (c) 4-bit Quantized MWD Encoded Map, (d) Red channel: sinusoidal encoding, $\sin(2\pi Z/P)$. (e) Green channel: sinusoidal encoding, $\cos(2\pi Z/P)$, (f) Blue channel: normalized depth for long-wavelength variation.}
\label{fig:visualize_mwd}
\end{figure}

\subsection{Multiwavelength Depth Encoding (MWD)}
We adopt the multiwavelength depth (MWD) transform proposed by Bell and Zhang \cite{bell2015multiwavelength}, as it enables conversion of a single-channel depth map $Z(i,j)$ into a smooth three-channel sinusoidal representation. This formulation enhances compressibility while preserving fine geometric detail. Let $P$ denote the high-frequency fringe period (in depth units); we use a fixed value of $P=8$ in all experiments for consistency. The red and green channels store wrapped phase via:
\begin{align}
I_{r}(i,j) &= 0.5[1.0+\sin(2\pi Z(i,j)/P)], \\
I_{g}(i,j) &= 0.5[1.0+\cos(2\pi Z(i,j)/P)].
\end{align}
To resolve phase ambiguities, the blue channel encodes a long-wavelength reference $R$:
\begin{equation}
I_{b}(i,j) = \frac{\phi_{b}(i,j)+\pi}{2\pi}
\end{equation}
where $\Phi(i,j)$ is derived from the depth $Z$. These three channels together produce the smooth sinusoidal patterns illustrated in Fig.~\ref{fig:visualize_mwd}, 
where \((I_r,I_g)\) capture high-frequency detail and \(I_b\) provides an absolute depth reference. 
Decoding first recovers the wrapped phase as
\begin{equation}
\phi(i,j) = \mathrm{atan2}\!\left(2I_r(i,j)-1,\;2I_g(i,j)-1\right),
\end{equation}
then extracts the fringe order \(k\) from \(I_b\), and finally computes the unwrapped phase
\begin{equation}
\Phi(i,j) = \phi(i,j) + 2\pi\,k.
\end{equation}
A linear mapping subsequently converts \(\Phi(i,j)\) back to the original depth map \(Z(i,j)\).
Integration with a learned compression backbone effectively leverages the smooth phase structure of the MWD representation. The sinusoidal encoding aligns well with the inductive priors of modern learned image codecs, while cross-channel trigonometric constraints stabilize quantization and decoding. In addition, the explicit geometric structure provided by MWD simplifies the learning problem by focusing the codec on residual redundancies. In our experiments, this integration reduces bitrate by 60.3\%, and the quantitative results are summarized in Section~4 (see Table~\ref{tab:middlebury}).

\subsection{Transformer-CNN Mixture Compression}
End-to-end learned image compression (LIC) has made rapid strides in recent years, often surpassing conventional codecs in rate-distortion performance. In DepthTCM, we adopt the Transformer-CNN mixture (TCM) hybrid autoencoder introduced by Liu et al. \cite{liu2023learned} as our compression backbone. This architecture comprises three main components:
\begin{enumerate}
    \item \textbf{Analysis transform $g_{a}$:} Alternating convolutional downsampling layers and Transformer-CNN Mixture (TCM) blocks encode the three-channel MWD input into a compact latent tensor $\mathbf{y}$.
    \item \textbf{Hyperprior module:} A lightweight network $h_{a}$ derives a side latent $\mathbf{z}$ from $\mathbf{y}$. We discretize and entropy-code $\mathbf{z}$, then use the decoded $\hat{\mathbf{z}}$ to parameterize the conditional entropy model for $\mathbf{y}$.
    \item \textbf{Synthesis transform $g_{s}$:} Mirroring $g_{a}$, inverse TCM blocks interleaved with upsampling convolutions reconstruct the quantized MWD image from the decoded latent $\hat{\mathbf{y}}$.
\end{enumerate}
All convolutional and self-attention weights are learned jointly under a rate-distortion objective, allowing DepthTCM to leverage both local texture priors and global context for efficient depth compression.

\subsection{Loss functions}
To train DepthTCM, we minimize a weighted multi-term loss that balances reconstruction fidelity, compression efficiency, and spatial regularization. The total objective is defined as:
\begin{equation}
\mathcal{L}_{total}=\lambda \cdot 255^{2}\mathcal{L}_{MSE}+\mathcal{L}_{Bpp}+\mathcal{L}_{Conf}+w_{TV}\mathcal{L}_{TV},
\end{equation}
where $\lambda$ controls rate-distortion trade-off and $w_{TV}=0.001$.

\textbf{Reconstruction loss ($\mathcal{L}_{MSE}$):}
\begin{equation}
\mathcal{L}_{MSE}=\frac{1}{HW}\sum{i,j}(\hat{Z}(i,j)-Z(i,j))^{2}
\end{equation}
Standard Mean-Squared Error (MSE) between reconstructed $\hat{Z}$ and ground-truth $Z$ depth is computed across all $H\times W$ pixels.

\textbf{Bitrate loss ($\mathcal{L}_{Bpp}$):}
\begin{equation}
\mathcal{L}_{Bpp} = \mathbb{E}[-\log_2 p(\mathbf{y}) - \log_2 p(\mathbf{z})]
\end{equation}
Negative log-likelihood of latents $\mathbf{y}$ and $\mathbf{z}$ under the learned entropy models \cite{balle2018variational, minnen2018joint}. Here, the expectation $\mathbb{E}$ is taken over the empirical distribution of the latents, directly minimizing the expected bits per pixel (Bpp).

\textbf{Confidence-weighted loss ($\mathcal{L}_{Conf}$):} Let $\Omega$ be the set of all pixels $(i,j)$ and $\tau\in(0,1)$ the relative threshold (we use 0.05). 
Define
\begin{equation}
T = \tau \max_{(i,j)\in\Omega}|\hat{Z}(i,j)-Z(i,j)|
\end{equation}
\begin{equation}
m(i,j) = \mathbf{1}(|\hat{Z}(i,j)-Z(i,j)| > T).
\end{equation}
Then the confidence-weighted loss is
\begin{equation}
\mathcal{L}_{Conf} = \frac{1}{|\Omega|} \sum_{(i,j) \in \Omega} m(i,j) (\hat{Z}(i,j) - Z(i,j))^2
\end{equation}

\textbf{Total Variation loss ($\mathcal{L}_{TV}$):}
\begin{equation}
\mathcal{L}_{\text{TV}} = \sum{i,j} \sqrt{ |\hat{Z}(i+1,j) - \hat{Z}(i,j)|^2 + |\hat{Z}(i,j+1) - \hat{Z}(i,j)|^2 }
\end{equation}
Encourages smoothness in homogeneous regions while retaining sharp depth edges. Each loss term serves a specific purpose: MSE for reconstruction accuracy, Bpp for compression efficiency, confidence-weighted loss for challenging regions, and TV for smoothness. Joint optimization enables DepthTCM to balance fidelity and bitrate, producing high-quality depth maps at low bitrates.

\section{Experiments}
\subsection{Experimental Setup}
\textbf{Implementation details.} We train DepthTCM on a subset of the ScanNet v2 dataset \cite{dai2017scannet}, sampling approximately 25,000 RGB-D frames from the training split to leverage its large scale, recency, and scene diversity (indoor layouts, sensors, and viewpoints), which are advantageous for generalization in real deployments. Unless stated otherwise, training runs for 100 epochs on an NVIDIA GeForce RTX 4080 SUPER with batch size 4; depth maps are prescaled for MWD compatibility.

\textbf{Benchmark and Test Scenarios.} We evaluate on six established benchmarks spanning different capture setups and resolutions: DIODE (indoor/outdoor RGB-D) \cite{vasiljevic2019diode}, Middlebury 2014 (high-resolution stereo) \cite{scharstein2014high}, ScanNet v2 test split \cite{dai2017scannet}, ScanNet++ iPhone RGB-D subset \cite{yeshwanth2023scannet++}, UnrealStereo4K dataset \cite{zhang2018unrealstereo}, and KITTI Depth Completion benchmark \cite{Uhrig2017THREEDV}. Test images are not used for optimization or model selection. For sparse-depth benchmarks such as KITTI Depth Completion, we compute losses and evaluation metrics only on valid depth pixels provided by the dataset (invalid pixels are ignored). Additional implementation specifics and extended qualitative results are provided in the supplementary material.

\textbf{Evaluation metrics.} We report bitrate (bits per pixel, bpp) measured on the entropy-coded stream and compression ratio (CR; original/compressed file sizes in KB). All bpp values are computed with respect to the original depth resolution ($H\times W$). Reconstruction quality is evaluated using PSNR, RMSE (root-mean-square error), and NRMSE (normalized RMSE), where NRMSE = RMSE $/(z_{max}-z_{min})$ with $z_{min}$ and $z_{max}$ denoting the minimum/maximum ground-truth depth. For readability, we additionally report a compact NRMSE-based percentage score, denoted as Accuracy:
\begin{equation}
\text{Accuracy} = \left( 1 - \frac{\text{RMSE}}{z_{max}-z_{min}} \right) \times 100\%.
\end{equation}
where $z_{min}$ and $z_{max}$ denote the minimum and maximum depth values in the ground-truth map, respectively. Note that this score is a monotonic transformation of NRMSE and is reported solely for interpretability; all primary comparisons are based on PSNR and RMSE. To assess inference speed, we measure the end-to-end compression and decompression runtimes.

\subsection{Rate-Distortion Performance}
DepthTCM demonstrates consistent improvements across benchmarks. On Middlebury 2014 (Table~\ref{tab:middlebury}), the 4-bit model achieves 49.89 dB PSNR at only 0.307 bpp, corresponding to a 44.29:1 compression ratio, outperforming both MWD (1.116 bpp, 28.66:1) and N-DEPTH (0.774 bpp, 41.35:1) while maintaining 99.38\% accuracy. Although the 8-bit variant achieves the highest PSNR (50.66 dB at 0.525 bpp), the 4-bit model provides the best overall trade-off between rate and distortion.

\begin{figure}[h]
  \centering
  \includegraphics[width=1.0\linewidth]{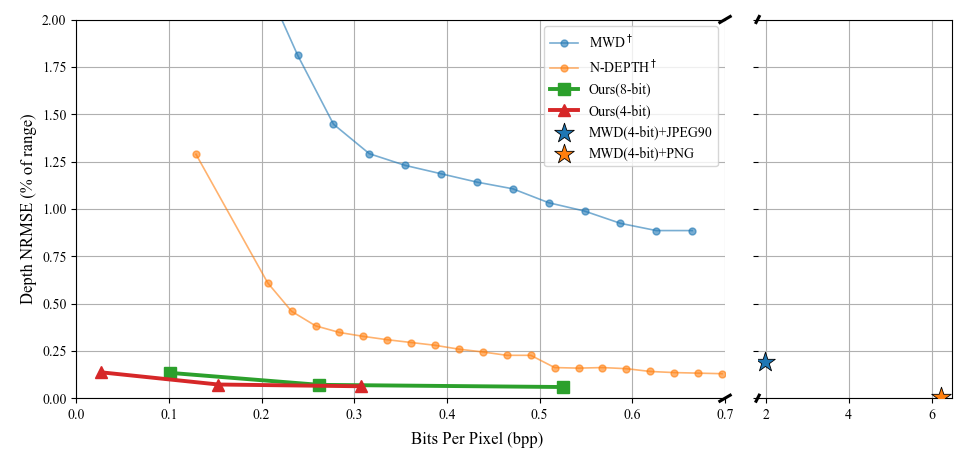}
  \caption{\textbf{Rate--distortion curves on the Middlebury 2014 dataset.} NRMSE (\%) is plotted against bitrate (bpp) measured per original depth pixel. The error is computed following the evaluation protocol of N-DEPTH, where metrics are calculated over the shared intersection of valid recovery regions. Rate--distortion operating points of our method are obtained by varying the Lagrangian multiplier $\lambda$.}
  \label{fig:rdcurve_diode}
\end{figure}

\begin{table}[t]
\centering
\caption{Quantitative comparison of depth compression methods on the Middlebury 2014 \cite{scharstein2014high} benchmark. Results marked with $\dagger$ are reported from the original publication\cite{siemonsma2024neural}, where JPEG(Q=90) is used as the entropy codec. Our method employs end-to-end learned entropy coding. All bitrate (bpp) values are computed per original depth pixel.}
\label{tab:middlebury}
\resizebox{\linewidth}{!}{
\begin{tabular}{l c c c c c c}
\toprule

Method & Codec & Bitrate$\downarrow$ (bpp) & PSNR$\uparrow$ (dB) & NRMSE$\downarrow$ (\%) & CR$\uparrow$ \\
\midrule
MWD$^\dagger$ & JPEG 90 & 1.116 & N/A & 1.012 & 28.66:1 \\
N-DEPTH$^\dagger$ & JPEG 90 & 0.774 & N/A & 0.068 & 41.35:1 \\
Ours (8-bit) & Learned & 0.525 & 50.66 & 0.059 & 26.39:1 \\
Ours (4-bit) & Learned & \textbf{0.307} & 49.89 & 0.063 & \textbf{44.29:1} \\
\bottomrule
\end{tabular}}
\end{table}

On DIODE, we analyze the proposed method through both RD curves (Fig.~\ref{fig:rdcurve_diode}) and quantitative comparisons (Table~\ref{tab:diode}). From the RD curves, obtained by varying the optimization parameter $\lambda$, the 4-bit model consistently provides a more favorable trade-off than the 8-bit variant, achieving comparable fidelity at substantially lower bitrates. Concretely, at the representative operating point ($\lambda=0.05$) the 4-bit DepthTCM reaches 44.31 dB at 0.365 bpp, whereas the 8-bit variant attains 44.99 dB at 1.075 bpp. This indicates that the 4-bit configuration is particularly effective in balancing rate and fidelity at low bitrates.

Table~\ref{tab:diode} further compares DepthTCM with conventional codecs, including the low-complexity lossless/near-lossless standard JPEG-LS. DepthTCM achieves significantly lower bitrates than JPEG 90 (0.365 vs. 2.191 bpp) while preserving much higher fidelity (44.31 dB vs. 26.87 dB). While JPEG-LS demonstrates high reconstruction quality (57.47 dB), its bitrate (4.287 bpp) is approximately $11\times$ higher than that of our 4-bit model. Even compared to PNG, our method operates at nearly $1/19$ the bitrate (0.365 vs. 6.896 bpp) while maintaining an accuracy of 99.84\%, which is highly comparable to the 99.86\% accuracy of JPEG-LS. These results highlight that DepthTCM offers a superior compression ratio for bandwidth-constrained applications without sacrificing essential geometric information.

\begin{table}[t]
\centering
\caption{Performance comparison with standard image codecs on the DIODE \cite{vasiljevic2019diode} dataset. Our method achieves a competitive accuracy comparable to the near-lossless JPEG-LS but at a significantly reduced bitrate.}
\label{tab:diode}
\resizebox{\linewidth}{!}{
\begin{tabular}{l c c c c c}
\toprule
Method & Codec & Bitrate$\downarrow$ (bpp) & PSNR$\uparrow$ (dB) & RMSE$\downarrow$ & Accuracy$\uparrow$ (\%) \\
\midrule
MWD (4-bit) & PNG       & 6.896 & 75.56 & 0.102 & 99.97 \\
MWD (4-bit) & JPEG 90 & 2.191 & 26.87 & 4.774 & 98.63 \\
MWD (4-bit) & JPEG-LS   & 4.287 & 57.47 & 0.340 & 99.86 \\
Ours (8-bit) & Learned & 1.075 & 44.99 & 0.509 & \textbf{99.85} \\
Ours (4-bit) & Learned & \textbf{0.363} & 44.31 & 0.535 & \textbf{99.84} \\
\bottomrule
\end{tabular}}
\end{table}

On the KITTI Depth Completion benchmark, which represents real-world outdoor driving scenarios with sparse LiDAR measurements, DepthTCM maintains a favorable balance between rate and fidelity (Table~\ref{tab:kitti}). At 0.580 bpp, the 4-bit model achieves an accuracy of 93.08\%, substantially reducing the bitrate compared to PNG (10.40 bpp) and JPEG 90 (4.401 bpp). While lossless PNG yields higher PSNR, it operates at a prohibitively high bitrate, whereas JPEG 90 exhibits severe degradation on sparse depth maps, as indicated by its low PSNR and large RMSE. Overall, these results indicate that DepthTCM remains robust across both dense and sparse depth distributions, bridging controlled benchmarks and real-world driving scenarios.
\begin{table}[t]
\centering
\caption{Performance comparison with standard image codecs on the KITTI Depth Completion benchmark \cite{Uhrig2017THREEDV}.}
\label{tab:kitti}
\resizebox{\linewidth}{!}{
\begin{tabular}{l c c c c c}
\toprule
Method & Codec & Bitrate$\downarrow$ (bpp) & PSNR$\uparrow$ (dB) & RMSE$\downarrow$ & Accuracy$\uparrow$ (\%) \\
\midrule
MWD (4-bit) & PNG      & 10.40 & 57.69 & 0.109 & 99.86 \\
MWD (4-bit) & JPEG 90  & 4.401 & 12.45 & 20.08 & 74.56 \\
Ours (8-bit) & Learned & 1.422 & 24.20 & 5.204 & \textbf{93.41} \\
Ours (4-bit) & Learned & \textbf{0.580} & 23.77 & 5.465 & 93.08 \\
\bottomrule
\end{tabular}}
\end{table}

On ScanNet v2 (Table~\ref{tab:scannet}), at substantially lower bitrates than JPEG 90, DepthTCM attains higher PSNR. Under aggressive compression, DepthTCM maintains PSNR in the high-50 dB range whereas JPEG 90 falls below 40 dB. Note that ScanNet v2 is an indoor dataset with a much narrower depth range compared to DIODE, which leads to larger differences in accuracy values despite similar RMSE. These results indicate that the codec adapts reliably across a wide range of target rates.

\begin{table}[t]
\centering
\caption{Performance comparison with standard image codecs on ScanNet v2 \cite{dai2017scannet}.}
\label{tab:scannet}
\resizebox{\linewidth}{!}{
\begin{tabular}{l c c c c c}
\toprule
Method & Codec & Bitrate$\downarrow$ (bpp) & PSNR$\uparrow$ (dB) & RMSE$\downarrow$ & Accuracy$\uparrow$ (\%) \\
\midrule
MWD (4-bit) & PNG      & 9.153 & 78.45 & 0.030 & 99.68 \\
MWD (4-bit) & JPEG 90  & 2.428 & 35.39 & 4.334 & 53.89 \\
Ours (8-bit) & Learned & 0.655 & 53.61 & 0.531 & \textbf{94.35} \\
Ours (4-bit) & Learned & \textbf{0.602} & 53.26 & 0.553 & \textbf{94.12} \\
\bottomrule
\end{tabular}}
\end{table}

On ScanNet++ iPhone RGB-D (Table~\ref{tab:scannetpp}), we evaluated a subset of 9,794 depth-only frames ($256\times192$) to test generalizability on recent low-resolution mobile LiDAR data. DepthTCM achieves 42.6--42.7 dB at 0.64--0.90 bpp with accuracy above 98.7\%, significantly outperforming JPEG 90 in both rate and fidelity. These results highlight the robustness of our approach across diverse resolutions and sensing modalities.

\begin{table}[t]
\centering
\caption{Performance comparison on the ScanNet++ \cite{yeshwanth2023scannet++} iPhone RGB-D subset.}
\label{tab:scannetpp}
\resizebox{\linewidth}{!}{
\begin{tabular}{l c c c c c}
\toprule
Method & Codec & Bitrate$\downarrow$ (bpp) & PSNR$\uparrow$ (dB) & RMSE$\downarrow$ & Accuracy$\uparrow$ (\%) \\
\midrule
MWD (4-bit) & PNG       & 11.528 & 78.19 & 0.013 & 99.98 \\
MWD (4-bit) & JPEG 90   & 2.554  & 34.68 & 0.086 & 97.09 \\
Ours (8-bit) & Learned  & 0.901  & 42.74 & 0.036 & \textbf{98.76} \\
Ours (4-bit) & Learned  & \textbf{0.638} & 42.63 & 0.037 & \textbf{98.75} \\
\bottomrule
\end{tabular}}
\end{table}

\begin{table}[!ht]
\centering
\caption{Performance comparison on the UnrealStereo4K dataset~\cite{zhang2018unrealstereo} (downsampled to $1920 \times 1080$ for testing).}
\label{tab:unrealstereo}
\resizebox{\linewidth}{!}{
\begin{tabular}{l c c c c c}
\toprule
Method & Codec & Bitrate$\downarrow$ (bpp) & PSNR$\uparrow$ (dB) & RMSE$\downarrow$ & Accuracy$\uparrow$ (\%) \\
\midrule
MWD (4-bit) & PNG       & 3.803 & 75.12 & 0.076 & 99.98 \\
MWD (4-bit) & JPEG 90   & 1.058 & 38.29 & 5.300 & 98.60 \\
Ours (8-bit) & Learned  & 0.543 & 49.14 & 1.653 & \textbf{99.57} \\
Ours (4-bit) & Learned  & \textbf{0.217} & 48.45 & 1.743 & \textbf{99.55} \\
\bottomrule
\end{tabular}}
\end{table}

Finally, to verify the scalability of DepthTCM on high-resolution data, we evaluated our method on the UnrealStereo4K dataset~\cite{zhang2018unrealstereo}. As shown in Table~\ref{tab:unrealstereo}, even when testing on Full HD resolution ($1920 \times 1080$), our 4-bit model achieves an extremely low bitrate of 0.217 bpp while maintaining a high PSNR of 48.45 dB. This is a significant improvement over JPEG 90, which requires nearly five times the bitrate (1.058 bpp) for a much lower quality (38.29 dB). These results demonstrate that our framework effectively handles larger spatial dimensions, preserving fine geometric details without incurring prohibitive bitrate costs.

\subsection{Runtime and Throughput}
Table~\ref{tab:inference} summarizes the average inference latency of our 4-bit model across various benchmarks. On the high-resolution DIODE dataset ($1024\times768$), our model achieves an encoding/decoding throughput of 6.08/5.13 FPS. As expected, performance scales with resolution, with throughput increasing to 12.08/10.37 FPS on ScanNet v2 ($640\times480$) and reaching near real-time speeds of 24.10/21.08 FPS on the lower-resolution ScanNet++ iPhone data. For a rough performance baseline, we compare our results to the published N-DEPTH \cite{siemonsma2024neural} 720p random-tensor reference (4.23/4.22 FPS). Although a direct comparison is challenging due to differing evaluation settings (real data vs. random tensors), our model demonstrates significantly higher throughput at comparable resolutions (DIODE vs. 720p) and scales efficiently to lower resolutions. Further details on hardware and numerical precision are provided in the supplementary material.

\begin{table}[t]
\centering
\caption{Average inference time (Ours, 4-bit) at benchmark input depth resolutions. Throughput is $1000/t$ (FPS).}
\label{tab:inference}
\resizebox{\linewidth}{!}{
\begin{tabular}{l c c c c}
\toprule
Dataset & Input & Enc. (ms) & Dec. (ms) & Enc./Dec. FPS \\
\midrule
DIODE & 1024x768 & 164.40 & 194.80 & 6.08 / 5.13 \\
ScanNet v2 & 640x480 & 82.78 & 96.46 & 12.08 / 10.37 \\
ScanNet++ & 256x192 & 41.48 & 47.45 & 24.10 / 21.08 \\
\midrule
N-DEPTH\cite{siemonsma2024neural} & 1280x720 & 236.40 & 237.20 & 4.23 / 4.22 \\
\bottomrule
\end{tabular}}
\end{table}

\subsection{Ablation Studies}

\begin{table}[t]
\centering
\caption{Ablation study of quantization bit-depths on the DIODE dataset.}
\label{tab:ablation_bit}
\begin{tabular}{l c c c c}
\toprule
Method & Bitrate$\downarrow$ (bpp) & PSNR$\uparrow$ (dB) & RMSE$\downarrow$ & Accuracy$\uparrow$ (\%) \\
\midrule
8-bit  & 1.075 & 44.99 & 0.509 & 99.21 \\
5-bit  & 0.397 & 44.61 & 0.544 & 99.16 \\
\textbf{4-bit} & \textbf{0.363} & \textbf{44.31} & \textbf{0.535} & \textbf{99.17} \\
3-bit  & 0.441 & 40.94 & 0.769 & 98.83 \\
2-bit  & 0.427 & 20.42 & 5.540 & 88.47 \\
\midrule
PNG    & 6.896 & 75.56 & 0.103 & 99.98 \\
JPEG   & 2.041 & 27.01 & 4.737 & 94.01 \\
\bottomrule
\end{tabular}
\end{table}

We conducted a series of ablation studies to validate the core components of DepthTCM: our choice of 4-bit quantization, the hybrid Transformer-CNN backbone, and our global quantization strategy. Our analysis confirmed that 4-bit quantization delivers optimal rate--distortion efficiency, achieving 44.31 dB PSNR on DIODE while striking the best balance across all tested bit-depths, as summarized in Table~\ref{tab:ablation_bit}. The hybrid Transformer-CNN backbone proved equally essential, delivering a PSNR gain of up to 0.75 dB over a CNN-only model at comparable bitrates (Table~\ref{tab:ablation_arch}). Collectively, these results confirm that the synergy between 4-bit encoding and an attention-based architecture consistently enhances rate-distortion performance.

\begin{table}[t]
\centering
\caption{Ablation study of architecture choices on Middlebury 2014 \cite{scharstein2014high}.}
\label{tab:ablation_arch}
\begin{tabular}{l c c c c}
\toprule
& \multicolumn{2}{c}{Transformer-CNN} & \multicolumn{2}{c}{CNN-only} \\
Method & PSNR & Bpp & PSNR & Bpp \\
\midrule
Ours (8-bit) & 50.66 & 0.525 & 49.91 & 0.532 \\
Ours (4-bit) & 49.89 & 0.307 & 49.22 & 0.259 \\
\bottomrule
\end{tabular}
\end{table}

\begin{table}[t]
\centering
\caption{Ablation study on quantization strategies.}
\label{tab:ablation_quant}
\begin{tabular}{l c c c c}
\toprule
Quantization & Bit Depth & PSNR & RMSE & Bitrate \\
\midrule
Adaptive & 2-6 bits & 47.82 & 0.018 & 0.412 \\
Fixed (Global) & 4 bits & 48.05 & 0.017 & 0.307 \\
\bottomrule
\end{tabular}
\end{table}
In addition to our proposed fixed global quantization, we tested an adaptive patch-wise quantization strategy where bit-depth was allocated based on local complexity. However, as shown in Table~\ref{tab:ablation_quant}, this adaptive approach proved less effective. The patch-wise processing introduced discontinuities at patch boundaries, disrupting the smooth sinusoidal patterns crucial for MWD encoding. This led to increased entropy in the latent representation and ultimately a higher bitrate for similar reconstruction quality. A detailed description of the adaptive quantization methodology is available in the supplementary material.

\subsubsection{Effectiveness of MWD Representation}
To address the lack of direct comparison with standard learned image compression (LIC) models adapted for depth, we established a baseline denoted as ``Only-TCM''. This baseline employs the TCM architecture~\cite{liu2023learned}, a state-of-the-art learned image compression model, trained directly on raw depth maps. This serves as a proxy for applying general-purpose learned 2D codecs (e.g., Cheng et al.~\cite{Cheng_2020_CVPR}, Minnen et al.~\cite{minnen2018joint}) to geometric data without domain-specific adaptation.

\begin{table}[t]
\centering
\caption{Comparison between our MWD-based approach and the standalone TCM backbone (Only-TCM) on the DIODE dataset. The ``Only-TCM'' baseline fails to capture geometric structures, resulting in high bitrates and low PSNR compared to our method.}
\label{tab:only_tcm}
\setlength{\aboverulesep}{0pt}
\setlength{\belowrulesep}{0pt}
\setlength{\extrarowheight}{.5ex}
\resizebox{\linewidth}{!}{
\begin{tabular}{l c c c c}
\toprule
Method & Bitrate$\downarrow$ (bpp) & PSNR$\uparrow$ (dB) & RMSE$\downarrow$ & Accuracy$\uparrow$ (\%) \\
\midrule
MWD (4-bit) + PNG & 6.896 & 75.56 & 0.102 & 99.97 \\
MWD (4-bit) + JPEG 90 & 2.191 & 26.87 & 4.774 & 98.63 \\
MWD (4-bit) + JPEG LS & 4.287 & 57.47 & 0.340 & 99.86 \\
\midrule
\textbf{Ours (8-bit)} & 1.075 & \textbf{44.99} & \textbf{0.509} & \textbf{99.85} \\
\textbf{Ours (4-bit)} & \textbf{0.363} & 44.31 & 0.535 & 99.84 \\
\midrule
Only-TCM & 1.443 & 22.38 & 1.213 & 92.42 \\
Only-TCM ($\lambda$ test) & 1.255 & 21.22 & 1.298 & 91.29 \\
Only-TCM (lr test) & 1.422 & 22.52 & 1.194 & 92.54 \\
\bottomrule
\end{tabular}}
\end{table}

As shown in Table~\ref{tab:only_tcm}, directly applying this 2D image compression model to depth data yields suboptimal performance. The standard ``Only-TCM'' model results in a high bitrate (1.443 bpp) with significantly lower PSNR (22.38 dB). We conducted two additional tests to investigate whether this failure stemmed from hyperparameter instability:
\begin{itemize}
\item \textbf{Effect of $\lambda$ Reduction:} We reduced the reconstruction weight $\lambda$ from 0.05 to 0.02. This successfully lowered the bitrate to 1.255 bpp but further degraded PSNR to 21.22 dB, confirming that the poor performance is not merely a trade-off issue.
\item \textbf{Effect of Learning Rate Decay:} To rule out optimization divergence, we decreased the learning rate from $1 \times 10^{-4}$ to $5 \times 10^{-5}$. While this slightly improved stability (22.52 dB PSNR), it failed to bridge the significant performance gap with our MWD-based method.
\end{itemize}

These results empirically demonstrate that simply retraining state-of-the-art 2D image compression models on depth data is insufficient due to the inherent statistical differences between texture and geometry. The degraded performance of Only-TCM can be attributed to the statistical properties of raw depth maps. Unlike natural images, depth maps exhibit large homogeneous regions separated by sharp discontinuities, leading to highly sparse and boundary-dominated residual patterns. These characteristics can be more challenging for general-purpose entropy models that are primarily optimized for natural image statistics. Without phase-structured reformulation, the learned hyperprior may struggle to model these distributions efficiently. In contrast, our MWD representation transforms depth into a spatially correlated RGB representation, enabling the TCM backbone to achieve 44.31 dB PSNR at a mere 0.363 bpp. This confirms that our framework effectively bridges the gap between geometric data and modern learned image codecs.

\subsubsection{Discussion}
\begin{table}[t]
\centering
\caption{Comparison with end-to-end depth compression methods. Direct metric comparison is omitted due to different datasets and evaluation protocols.}
\label{tab:e2e_depth_comparison}
\small
\setlength{\tabcolsep}{4pt}
\renewcommand{\arraystretch}{1.2}
\resizebox{\linewidth}{!}{%
\begin{tabular}{@{} l c c l p{6.0cm} @{}}
\toprule
Method & Inputs & Rate Control & Architecture Features & Notes \\
\midrule
Texture-Guided DMC~\cite{peng2022texture}
& Depth+RGB & Fixed & $\bullet$ RGB guidance
& $\bullet$ Requires paired RGB (not always available). \\

Lossless High-Precision~\cite{wu2022end}
& Depth only & N/A & $\bullet$ Pseudo-residual
& $\bullet$ Focused only on archival lossless coding (not RD trade-offs). \\

Variable-Rate DMC~\cite{sebai2024end}
& Depth only & Variable ($\lambda$)
& \begin{tabular}[t]{@{}l@{}} $\bullet$ Wedgelet filters \\ $\bullet$ VGG19 encoder \end{tabular}
& \begin{tabular}[t]{@{}p{6.0cm}@{}} $\bullet$ Synthesized-view RD (original texture + compressed depth); limited cross-dataset reproducibility. \end{tabular} \\

\textbf{DepthTCM (Ours)}
& \textbf{Depth only} & \textbf{Variable ($\lambda$)}
& \begin{tabular}[t]{@{}l@{}} \textbf{$\bullet$ Physics-based MWD} \\ \textbf{$\bullet$ Transformer--CNN} \end{tabular}
& \begin{tabular}[t]{@{}p{6.0cm}@{}} \textbf{$\bullet$ Depth-only; physics-based transform improves RD efficiency.} \end{tabular} \\
\bottomrule
\end{tabular}}
\end{table}

Table~\ref{tab:e2e_depth_comparison} contrasts inputs, rate control, and architectural priors across recent end-to-end approaches. Texture-guided DMC~\cite{peng2022texture} requires paired RGB and therefore depends on an additional sensing modality at deployment. By contrast, the lossless high-precision method~\cite{wu2022end} is aimed at archival compression and does not provide rate--distortion trade-offs. The variable-rate approach by Sebai et al.~\cite{sebai2024end} offers single-model rate control via the parameter~$\lambda$, but its evaluation focuses on synthesized-view RD, which can limit cross-dataset reproducibility. For instance, their method operates at a very low bitrate regime (below 0.1~bpp) on datasets such as Ballet and Champagne, reaching a peak PSNR in the 38--45~dB range. In contrast, our proposed DepthTCM achieves a significantly higher PSNR of 49.89~dB at 0.307~bpp on the Middlebury 2014 benchmark. Although differing evaluation settings preclude a direct comparison, this result indicates that DepthTCM is optimized for a higher-fidelity operating point than methods focused on ultra-low bitrates.

In contrast to these methods, DepthTCM operates on depth only, supports variable-rate coding via $\lambda$, and combines a physics-based MWD pre-transform with a Transformer--CNN backbone. This design yields entropy-friendly latents and stable reconstruction without auxiliary modalities, which aligns with the empirical RD gains observed on DIODE, Middlebury~2014, and ScanNet~v2 (Tables~\ref{tab:diode}--\ref{tab:scannet} and Fig.~\ref{fig:rdcurve_diode}).

\section{Conclusion}
In this paper, we introduced DepthTCM, a novel end-to-end framework for depth map compression that synergizes a physics-inspired multiwavelength depth (MWD) encoding with a hybrid Transformer-CNN mixture architecture. Our core contribution lies in demonstrating that mapping depth to a 4-bit quantized sinusoidal representation drastically lowers entropy while preserving high geometric fidelity. Extensive experiments validate that DepthTCM surpasses both traditional codecs and prior multiwavelength methods in rate-distortion performance across several standard benchmarks. This work highlights a promising direction for geometric data compression through the synergy of domain-specific priors and modern deep learning architectures. Future work will address limitations, such as performance in highly dynamic scenes, by extending the framework to temporal depth-video compression and optimizing it for real-time embedded systems to improve robustness and broaden applicability.

\section*{CRediT authorship contribution statement}
\textbf{Young-Seo Chang:} Writing – original draft, Validation, Methodology, Formal analysis.
\textbf{Yatong An:} Writing – review \& editing, Validation, Funding acquisition.
\textbf{Jae-Sang Hyun:} Writing – review \& editing, Validation, Supervision, Project administration, Funding acquisition.

\section*{Declaration of Competing Interest}
The authors declare that they have no known competing financial interests or personal relationships that could have appeared to influence the work reported in this paper.

\section*{Data Availability}
All experiments were conducted on publicly available datasets. The code and trained models that support the findings of this study are available from the corresponding author upon reasonable request.

\section*{Acknowledgements}
This research was supported by the Technology Innovation Program(Project Name: Development of AI autonomous continuous production system technology for gas turbine blade maintenance and regeneration for power generation, Project Number: RS-2025-25447257, Contribution Rate: 50\%) funded By the Ministry of Trade, Industry and Resources(MOTIR, Korea), the Culture, Sports and Tourism R\&D Program through the Korea Creative Content Agency grant funded by the Ministry of Culture, Sports and Tourism in 2024 (Project Name: Global Talent for Generative AI Copyright Infringement and Copyright Theft, Project Number: RS-2024- 00398413, Contribution Rate: 40\%), and the National Research Foundation of Korea(NRF) grant funded by the Korea government(MSIT) (Project Number: RS-2025-16072782, Contribution Rate: 10\%)
%% References
\bibliographystyle{elsarticle-num} 
\bibliography{ref}

\end{document}